\documentclass{article}

 \usepackage[main, final]{neurips_2025}
\usepackage[utf8]{inputenc} % allow utf-8 input
\usepackage[T1]{fontenc}    % use 8-bit T1 fonts
\usepackage{hyperref}       % hyperlinks
\usepackage{url}            % simple URL typesetting
\usepackage{booktabs}       % professional-quality tables
\usepackage{amsfonts}       % blackboard math symbols
\usepackage{nicefrac}       % compact symbols for 1/2, etc.
\usepackage{microtype}      % microtypography
\usepackage{xcolor}         % colors
\usepackage{natbib}
\usepackage{amsmath}
\usepackage{graphicx}

\usepackage[capitalise,nameinlink]{cleveref}
\usepackage{booktabs,wrapfig}
\usepackage{tcolorbox}
\tcbuselibrary{listings}

\title{Physics-Informed Large Language Models for HVAC Anomaly Detection with Autonomous Rule Generation}

\author{%
  Subin Lin\\
  Department of the Built Environment\\
  National University of Singapore\\
  \texttt{subin\_lin@u.nus.edu}\\
  \And
  Chuanbo Hua\\
  InnoCore PRISM-AI\\
  KAIST\\
  \texttt{cbhua@kaist.ac.kr}\\
}

\begin{document}

\maketitle

\begin{abstract}
Heating, Ventilation, and Air-Conditioning (HVAC) systems account for a substantial share of global building energy use, making reliable anomaly detection essential for improving efficiency and reducing emissions. Classical rule-based approaches offer explainability but lack adaptability, while deep learning methods provide predictive power at the cost of transparency, efficiency, and physical plausibility. Recent attempts to use Large Language Models (LLMs) for anomaly detection improve interpretability but largely ignore the physical principles that govern HVAC operations. We present PILLM, a \textbf{P}hysics-\textbf{I}nformed \textbf{LLM} framework that operates within an evolutionary loop to automatically generate, evaluate, and refine anomaly detection rules. Our approach introduces physics-informed reflection and crossover operators that embed thermodynamic and control-theoretic constraints, enabling rules that are both adaptive and physically grounded. Experiments on the public Building Fault Detection dataset show that PILLM achieves state-of-the-art performance while producing diagnostic rules that are interpretable and actionable, advancing trustworthy and deployable AI for smart building systems.
\end{abstract}

\section{Introduction}
The global imperative to mitigate climate change has placed the urban built environment at the forefront of sustainability research. Buildings account for approximately 40\% of global energy consumption and a third of greenhouse gas emissions, making them a critical leverage point for decarbonization \citep{UNEP2021}. The complex Heating, Ventilation, and Air-Conditioning (HVAC) systems within them are major consumers of this energy. However, anomalies in HVAC system operation not only undermine energy efficiency but are also difficult to detect amidst the complexity and scale of building data, underscoring the critical need for robust anomaly detection methods \citep{Amasyali2018}.

% While modern building management systems generate vast streams of operational data, a persistent and well-documented gap remains between this wealth of data and the ability to derive the actionable intelligence required to optimize performance .\\

Automated Fault Detection and Diagnostics (AFDD) has long been pursued to address anomalies in HVAC systems. 
Recent work emphasizes that effective anomaly detection must jointly satisfy \textit{explainability}, \textit{reproducibility}, and \textit{autonomy}. Classical rule-based methods can detect explainable predefined faults \citep{Katipamula2005}, but they require expert-crafted knowledge, are static in the face of evolving building dynamics, and struggle with the complexity of real-world operations \citep{kim2018review}. Deep learning methods, including LSTM and Transformer-based architectures, have since shown strong predictive performance by uncovering subtle, non-linear patterns \citep{karpontinis2024transformer, Wang2020}. However, they remain difficult to deploy in practice: models often act as black boxes, demand heavy computation, and generalize poorly when physical knowledge of the built environment is not incorporated \citep{jiang2024modularized}. These trade-offs highlight a persistent tension between the interpretability of heuristics and the accuracy.

Recently, Large Language Models (LLMs) have emerged as a promising tool for rule design in anomaly detection. By generating human-readable heuristics and providing natural-language rationales, LLM-based methods enhance explainability and reduce the manual effort required for rule construction \citep{liu2025large, ye2024reevo, Lin2025BuildEvo}. However, current LLM-based approaches often overlook critical physical constraints and domain knowledge inherent to HVAC systems. Without grounding anomaly detection in these real-world physical principles, the resulting rules risk being incomplete, misaligned with building dynamics, or prone to false alarms. Bridging LLM-driven rule generation with physically grounded knowledge therefore represents a crucial step toward developing anomaly detection systems that are not only explainable and adaptive, but also robust and trustworthy in practical deployment.

To address the limitations of prior approaches, we present Physics-Informed Large Language Model (PILLM), a framework wherein LLMs operate within an evolutionary loop to automatically generate, evaluate, and refine anomaly detection rules, critically guided by real-world physical principles to ensure transparency and plausibility. Our approach automatically incorporates real-world physical principles into the rule generation process. By combining LLMs’ world knowledge with curated building context and sensor data, PILLM generates diagnostic rules that are both transparent and physically plausible. Furthermore, we embed physical constraints directly into the evolutionary optimization process through novel reflection and crossover operators, ensuring that the generated rules remain aligned with thermodynamic and control-theoretic principles. 

Our main contributions are as follows:
\begin{enumerate}
\item We propose PILLM, a novel framework that integrates LLMs with evolutionary search to automatically generate anomaly detection rules while explicitly incorporating building physics and operational semantics. 
\item We design physics-informed reflection and crossover mechanisms that guide LLM-generated rules toward physical plausibility and robustness, addressing the limitations of purely statistical or heuristic-based approaches. 
\item We evaluate our framework on the public LBNL Automated Fault Detection for Buildings dataset, showing that it achieves state-of-the-art performance while producing interpretable and actionable diagnostic rules. 
\end{enumerate}

\section{Related Work}

\textbf{LLM for Anomaly Detection} A systematic literature review highlights that LLMs can serve three main roles: augmenting detection pipelines with synthetic data or pseudo-labels, acting directly as anomaly/out-of-distribution detectors, and generating interpretable explanations for detection outcomes \citep{liu2025large}. In time-series settings, methods like LLMAD employ retrieval of similar patterns and a chain-of-thought reasoning strategy to deliver both accurate and interpretable results \citep{liu2025large}. SigLLM further explores dual operational modes for time-series anomaly detection: in \textit{Detector mode}, LLMs predict the next steps in the sequence and identify anomalies by comparing predictions with ground-truth signals, while in prompter mode, LLMs are directly prompted with time-series data to localize anomalous indices \citep{alnegheimish2024large}. Other systems adopt an agentic paradigm, for instance, Argos uses LLMs to autonomously generate explainable anomaly rules in an iterative, rule-based framework, achieving significant accuracy improvements \citep{gu2025argos}. In the specific context of building HVAC systems, LLMs such as DistilBERT have been fine-tuned to classify operational fault conditions from time-series data, demonstrating strong performance (F1 scores up to 99\%) and robustness to noisy inputs \citep{langer2024large}. These developments underscore the flexibility of LLMs in anomaly detection tasks, particularly for enhancing explainability, adaptability, and performance across varied application domains.

Further references on classical approaches and deep learning methods can be found in the appendix.

% \section{Why HVAC Datasets are Ideal for Physics-Informed Models}
% The selection of an HVAC operational dataset for this work is deliberate, as these cyber-physical systems represent a distinct and highly suitable domain for the application of physics-informed models. Unlike domains such as natural language processing or computer vision, where underlying patterns are statistical and emergent, the behavior of an HVAC system is a direct manifestation of causal relationships dictated by the well-understood first principles of thermodynamics, fluid dynamics, and heat transfer. A purely data-driven model, agnostic to these principles, is inherently susceptible to learning spurious, non-physical correlations from the training data, leading to poor generalization and a failure to build trustworthy systems \citep{willard2022integrating}. The known physical laws governing HVAC operation provide a powerful inductive bias, constraining a model's hypothesis space to physically plausible solutions. This approach is critical for enhancing model robustness and generalizability, particularly in the presence of sensor noise or novel operational scenarios. Therefore, this domain serves as an ideal and rigorous testbed for developing and validating the capabilities of a physics-informed reasoning framework like PILLM, where the ultimate goal is not just pattern recognition, but achieving a human-like, causal understanding of a complex engineered system.
\section{Methodology}

\begin{figure}[t]
  \centering
  \includegraphics[scale=1.05]{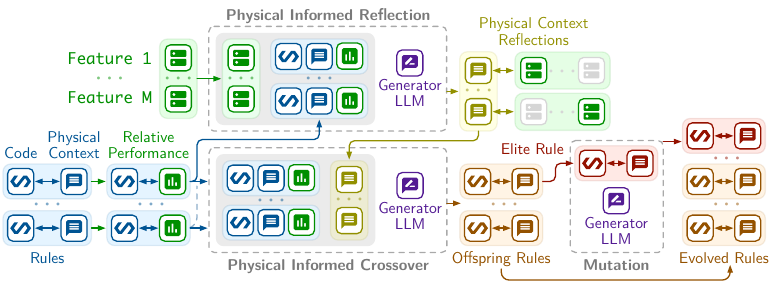}
  \caption {Overview of PILLM. The framework follows an evolutionary generate-and-reflect pipeline for anomaly detection rules. In each iteration, the current rule population undergoes \textit{Physical-Informed Reflection}, where physical context is incorporated into candidate rules. These reflections are then used in \textit{Physical-Informed Crossover} to produce the next generation of rules. Finally, elite rules are refined through mutation, resulting in evolved rules that are adaptive.}
  \label{fig:overview}
\end{figure}

In this section, we present PILLM as illustrated in \cref{fig:overview}. We introduce two key components : \textit{Physical Informed Reflection} (PIR), and \textit{Physical Informed Crossover} (PIC). Together with the evolving anomaly detection rules generation pipeline, these components enable dynamic, flexible, and smart way to embed the physical information into the rule generation. We then lay out the components details and the training scheme.

% \subsection{Dataset and Preprocessing}

\subsection{PILLM}

Our framework builds on the Reflective Evolution paradigm \citep{ye2024reevo}, where LLMs are employed as reasoning engines to perform genetic operators—initialization, reflection, crossover, and mutation—while being explicitly guided by physical knowledge of HVAC systems. Unlike conventional evolutionary approaches, PILLM does not treat heuristics as abstract code snippets. Instead, each rule is continuously contextualized by its physical meaning (e.g., temperature dynamics, airflow, occupancy schedules), ensuring that the evolutionary process remains grounded in real-world building physics.

\paragraph{Initial Population.}
The process begins by prompting the generator LLM with a task specification for anomaly detection rules. The specification defines the inputs (e.g., room and floor temperatures, fan status, fan speed), the output (an anomaly score), and the objective function (e.g., maximize detection accuracy). To seed the process, the LLM is also provided with a simple baseline heuristic (e.g., a peak-over-threshold rule). From this prompt, the LLM generates a diverse population of $N$ initial rule candidates in executable code form, each accompanied by a short natural-language rationale. This ensures diversity not only in implementation but also in interpretability.

\paragraph{Physics-Informed Reflection.}
At each iteration, candidate rules are reflected upon using physical context. The reflection stage compares the relative performance of rules and analyzes their alignment with the real-world meaning of input features. Crucially, the LLM is provided with metadata describing each feature’s physical role in the HVAC system (e.g., “Zone temperature reflects indoor thermal conditions,” “Fan speed governs airflow rate and pressure”). The LLM then produces structured reflections that highlight which physical aspects a rule captures and which are neglected. For example, a reflection might conclude that a rule focusing exclusively on outdoor temperature misses critical dynamics of indoor load variation. These reflections serve as a bridge between raw performance metrics and domain knowledge, guiding the evolutionary process toward rules that are both effective and physically sound.

\paragraph{Physics-Informed Crossover.}
Reflections directly shape the crossover operation. Instead of combining rules blindly, the LLM merges parent rules in a way that respects and integrates their associated physical contexts. For instance, one parent rule may emphasize temperature fluctuations across indoor and outdoor sensors, while another focuses on fan speed and airflow pressure. Through physics-informed crossover, the offspring rule may learn to model the causal relationship between thermal gradients and airflow control, yielding a more coherent and actionable heuristic. By explicitly anchoring code recombination to physical interpretations, this stage avoids the generation of arbitrary hybrids and instead synthesizes offspring with meaningful improvements in diagnostic coverage.

\paragraph{Elitist Rule Mutation.}
Finally, elite rules undergo mutation guided by long-term reflections. Instead of wholesale rewrites, the LLM proposes targeted refinements, such as adding occupancy schedules or weather normalization, to enhance robustness and generalizability.

\section{Experiment}

For more details about dataset preprocessing, hyperparameters, baseline settings, hardware and software environment, as well as additional results and analysis, please refer to the appendix. 

\begin{wraptable}{r}{0.5\textwidth}
\vspace{-\baselineskip}
\caption{Performance results of different anomaly detection baselines. Best and second best results are in \textbf{bold} and \underline{underline}.}
\centering
\small
\setlength{\tabcolsep}{6pt}
\renewcommand{\arraystretch}{1.15}
\begin{tabular}{lccc}
\toprule
\textbf{Method} & \textbf{Precision} & \textbf{Recall} & $\mathbf{F_{1}}$ \\
\midrule
AnomalyTransformer   & 0.482 & 0.395 & 0.282 \\
AutoRegression       & 0.731 & 0.699 & 0.668 \\
LSTMAD               & 0.861 & 0.781 & 0.818 \\
LLMAD                & 0.045 & 0.835 & 0.083 \\
SigLLM               & 0.012 & 0.502 & 0.021 \\
ARGOS                & 0.921 & \textbf{0.885} & \underline{0.902} \\
\midrule
PILLM                & \textbf{0.968} & \underline{0.859} & \textbf{0.926} \\
\quad w/o PIR        & 0.889 & 0.851 & 0.869 \\
\quad w/o PIC        & \underline{0.945} & 0.803 & 0.868 \\
\bottomrule
\end{tabular}
\label{tab:kpi-results}
\vspace{-5mm}
\end{wraptable}

\paragraph{Main Results.}
We report the performance of PILLM against a set of benchmark methods in \cref{tab:kpi-results}. Across all baselines, PILLM achieves the highest precision and $F_{1}$ score, while maintaining competitive recall. In particular, ARGOS achieves the strongest recall, but its overall performance remains slightly below PILLM in terms of $F_{1}$. Other classical (e.g., AutoRegression, LSTMAD) and LLM-based baselines (e.g., LLMAD, SigLLM) lag behind, reflecting either limited adaptability or poor precision. These results confirm that PILLM not only produces state-of-the-art performance but also balances accuracy with physical plausibility. 

\paragraph{Ablation Study.}
We further analyze the role of physics-informed components by ablating PIR and PIC. As shown in \cref{tab:kpi-results}, removing either PIR or PIC leads to clear performance degradation, particularly in $F_{1}$. Without PIR, the model underperforms in aligning rules with feature semantics, while without PIC, the offspring rules become less coherent and lose physical grounding. These results validate the importance of explicitly embedding physical knowledge in the evolutionary loop. 

\paragraph{Explainability.}
A key advantage of PILLM is that it generates anomaly detection rules in executable, human-readable Python code. Unlike neural baselines that act as black boxes, the heuristics evolved by PILLM are transparent and easily interpretable. For example, an evolved rule might explicitly check for abnormal thermal gradients in relation to fan speed or weather conditions, providing clear physical reasoning behind the anomaly flag. This interpretability enhances trust and usability for building operators, who can validate, debug, and refine the generated rules with domain expertise. By producing rules that are both performant and understandable, PILLM bridges the gap between machine learning advances and real-world operational deployment.

\section{Conclusion}
In this work, we introduced PILLM, a physics-informed LLM framework for anomaly detection in HVAC systems. By embedding domain knowledge into the evolutionary generation of rules through physics-informed reflection and crossover, PILLM bridges the gap between adaptability and physical plausibility. Experiments on the LBNL Automated Fault Detection dataset demonstrate that PILLM achieves state-of-the-art precision and $F_{1}$ score while maintaining competitive recall, outperforming both classical and neural baselines. Beyond accuracy, PILLM produces rules that are interpretable and actionable, offering building operators transparent insights into system faults. These results highlight the promise of combining LLM reasoning with physics-informed optimization to advance trustworthy and deployable AI for cyber-physical systems. Future work will explore extending PILLM to other building subsystems and investigating its scalability to real-time anomaly detection in large-scale smart infrastructure.

\clearpage

\bibliographystyle{plainnat}
\bibliography{reference}

\begin{thebibliography}{21}
\providecommand{\natexlab}[1]{#1}
\providecommand{\url}[1]{\texttt{#1}}
\expandafter\ifx\csname urlstyle\endcsname\relax
  \providecommand{\doi}[1]{doi: #1}\else
  \providecommand{\doi}{doi: \begingroup \urlstyle{rm}\Url}\fi

\bibitem[Alnegheimish et~al.(2024)Alnegheimish, Nguyen, Berti-Equille, and Veeramachaneni]{alnegheimish2024large}
Sarah Alnegheimish, Linh Nguyen, Laure Berti-Equille, and Kalyan Veeramachaneni.
\newblock Large language models can be zero-shot anomaly detectors for time series?
\newblock \emph{arXiv preprint arXiv:2405.14755}, 2024.

\bibitem[Amasyali and El-Gohary(2018)]{Amasyali2018}
K.~Amasyali and N.~M. El-Gohary.
\newblock A review of data-driven building energy consumption prediction studies.
\newblock \emph{Renewable and Sustainable Energy Reviews}, 81:\penalty0 1192--1205, 2018.
\newblock \doi{10.1016/j.rser.2017.04.067}.

\bibitem[Ciobanu-Caraus et~al.(2024)Ciobanu-Caraus, Aicher, Kernbach, Regli, Serra, and Staartjes]{ciobanu2024critical}
Olga Ciobanu-Caraus, Anatol Aicher, Julius~M Kernbach, Luca Regli, Carlo Serra, and Victor~E Staartjes.
\newblock A critical moment in machine learning in medicine: on reproducible and interpretable learning.
\newblock \emph{Acta neurochirurgica}, 166\penalty0 (1):\penalty0 14, 2024.

\bibitem[Comanici et~al.(2025)Comanici, Bieber, Schaekermann, Pasupat, Sachdeva, Dhillon, Blistein, Ram, Zhang, Rosen, et~al.]{comanici2025gemini}
Gheorghe Comanici, Eric Bieber, Mike Schaekermann, Ice Pasupat, Noveen Sachdeva, Inderjit Dhillon, Marcel Blistein, Ori Ram, Dan Zhang, Evan Rosen, et~al.
\newblock Gemini 2.5: Pushing the frontier with advanced reasoning, multimodality, long context, and next generation agentic capabilities.
\newblock \emph{arXiv preprint arXiv:2507.06261}, 2025.

\bibitem[Cuomo et~al.(2022)Cuomo, Di~Cola, Giampaolo, Rozza, Raissi, and Piccialli]{cuomo2022scientific}
Salvatore Cuomo, Vincenzo~Schiano Di~Cola, Fabio Giampaolo, Gianluigi Rozza, Maziar Raissi, and Francesco Piccialli.
\newblock Scientific machine learning through physics--informed neural networks: Where we are and what’s next.
\newblock \emph{Journal of Scientific Computing}, 92\penalty0 (3):\penalty0 88, 2022.

\bibitem[Gu et~al.(2025)Gu, Xiong, Mace, Jiang, Hu, Kasikci, and Cheng]{gu2025argos}
Yile Gu, Yifan Xiong, Jonathan Mace, Yuting Jiang, Yigong Hu, Baris Kasikci, and Peng Cheng.
\newblock Argos: Agentic time-series anomaly detection with autonomous rule generation via large language models.
\newblock \emph{arXiv preprint arXiv:2501.14170}, 2025.

\bibitem[Jiang and Dong(2024)]{jiang2024modularized}
Zixin Jiang and Bing Dong.
\newblock Modularized neural network incorporating physical priors for future building energy modeling.
\newblock \emph{Patterns}, 5\penalty0 (8), 2024.

\bibitem[Karpontinis and Alexandridis(2024)]{karpontinis2024transformer}
Dimitrios Karpontinis and Georgios Alexandridis.
\newblock Transformer-based anomaly detection in energy consumption data.
\newblock In \emph{IFIP International Conference on Artificial Intelligence Applications and Innovations}, pages 325--331. Springer, 2024.

\bibitem[Katipamula and Brambley(2005)]{Katipamula2005}
S.~Katipamula and M.~R. Brambley.
\newblock Methods for fault detection, diagnostics, and prognostics for building systems---a review, part i.
\newblock \emph{HVAC\&R Research}, 11\penalty0 (1):\penalty0 3--25, 2005.
\newblock \doi{10.1080/10789669.2005.10391108}.

\bibitem[Kim and Katipamula(2018)]{kim2018review}
Woohyun Kim and Srinivas Katipamula.
\newblock A review of fault detection and diagnostics methods for building systems.
\newblock \emph{Science and Technology for the Built Environment}, 24\penalty0 (1):\penalty0 3--21, 2018.

\bibitem[Kojima et~al.(2022)Kojima, Gu, Reid, Matsuo, and Iwasawa]{kojima2022large}
Takeshi Kojima, Shixiang~Shane Gu, Machel Reid, Yutaka Matsuo, and Yusuke Iwasawa.
\newblock Large language models are zero-shot reasoners.
\newblock \emph{Advances in neural information processing systems}, 35:\penalty0 22199--22213, 2022.

\bibitem[Langer et~al.(2024)Langer, Hirsch, Kern, Kohl, and Schweiger]{langer2024large}
Gerda Langer, Thomas Hirsch, Roman Kern, Theresa Kohl, and Gerald Schweiger.
\newblock Large language models for fault detection in buildings’ hvac systems.
\newblock In \emph{Energy Informatics Academy Conference}, pages 49--60. Springer, 2024.

\bibitem[Lin and Hua(2025)]{Lin2025BuildEvo}
Subin Lin and Chuanbo Hua.
\newblock Buildevo: Designing building energy consumption forecasting heuristics via llm-driven evolution.
\newblock \emph{arXiv preprint arXiv:2507.12207}, 2025.
\newblock URL \url{https://arxiv.org/abs/2507.12207}.

\bibitem[Liu et~al.(2025)Liu, Zhang, Qian, Ma, Qin, Bansal, Lin, Rajmohan, and Zhang]{liu2025large}
Jun Liu, Chaoyun Zhang, Jiaxu Qian, Minghua Ma, Si~Qin, Chetan Bansal, Qingwei Lin, Saravan Rajmohan, and Dongmei Zhang.
\newblock Large language models can deliver accurate and interpretable time series anomaly detection.
\newblock In \emph{Proceedings of the 31st ACM SIGKDD Conference on Knowledge Discovery and Data Mining V. 2}, pages 4623--4634, 2025.

\bibitem[Raissi et~al.(2019)Raissi, Perdikaris, and Karniadakis]{raissi2019physics}
Maziar Raissi, Paris Perdikaris, and George~E Karniadakis.
\newblock Physics-informed neural networks: A deep learning framework for solving forward and inverse problems involving nonlinear partial differential equations.
\newblock \emph{Journal of Computational physics}, 378:\penalty0 686--707, 2019.

\bibitem[{United Nations Environment Programme}(2021)]{UNEP2021}
{United Nations Environment Programme}.
\newblock 2021 global status report for buildings and construction: Towards a zero-emission, efficient and resilient buildings and construction sector.
\newblock Technical report, UNEP, Nairobi, 2021.

\bibitem[Wang et~al.(2020)Wang, Wang, Hong, and Piette]{Wang2020}
Z.~Wang, K.~Wang, T.~Hong, and M.~Piette.
\newblock A novel methodology for creating scheduled and unscheduled building occupancy data.
\newblock \emph{Energy and Buildings}, 223:\penalty0 110196, 2020.
\newblock \doi{10.1016/j.enbuild.2020.110196}.

\bibitem[Wei et~al.(2022)Wei, Wang, Schuurmans, Bosma, Xia, Chi, Le, Zhou, et~al.]{wei2022chain}
Jason Wei, Xuezhi Wang, Dale Schuurmans, Maarten Bosma, Fei Xia, Ed~Chi, Quoc~V Le, Denny Zhou, et~al.
\newblock Chain-of-thought prompting elicits reasoning in large language models.
\newblock \emph{Advances in neural information processing systems}, 35:\penalty0 24824--24837, 2022.

\bibitem[Ye et~al.(2024)Ye, Wang, Cao, Berto, Hua, Kim, Park, and Song]{ye2024reevo}
Haoran Ye, Jiarui Wang, Zhiguang Cao, Federico Berto, Chuanbo Hua, Haeyeon Kim, Jinkyoo Park, and Guojie Song.
\newblock Reevo: Large language models as hyper-heuristics with reflective evolution.
\newblock \emph{Advances in neural information processing systems}, 37:\penalty0 43571--43608, 2024.

\bibitem[Zhang et~al.(2023)Zhang, Saeed, and Sadeghian]{zhang2023deep}
Fan Zhang, Nausheen Saeed, and Paria Sadeghian.
\newblock Deep learning in fault detection and diagnosis of building hvac systems: A systematic review with meta analysis.
\newblock \emph{Energy and AI}, 12:\penalty0 100235, 2023.

\bibitem[Zhao et~al.(2019)Zhao, Li, Zhang, and Zhang]{zhao2019artificial}
Yang Zhao, Tingting Li, Xuejun Zhang, and Chaobo Zhang.
\newblock Artificial intelligence-based fault detection and diagnosis methods for building energy systems: Advantages, challenges and the future.
\newblock \emph{Renewable and Sustainable Energy Reviews}, 109:\penalty0 85--101, 2019.

\end{thebibliography}

\clearpage

\section*{Appendix}

\subsection*{Detailed Problem Definition}
\paragraph{Task}
We address building-level anomaly detection in HVAC systems using multivariate time-series data. 
Given a building \(b\) with sensor set \(\mathcal{F}_b = \{f_1, f_2, \dots, f_M\}\), the input at each timestep \(t\) is a feature vector 
\(\mathcal{S}_b^t = (x_{b,f_1}^t, x_{b,f_2}^t, \dots, x_{b,f_M}^t)\), 
where \(x_{b,f}^t \in \mathbb{R}\) denotes the reading of feature \(f\) (e.g., zone temperature, fan speed, air flow rate). 
The goal is to learn a mapping from the observed sequence 
\(H_b = (\mathcal{S}_b^1, \dots, \mathcal{S}_b^{T_{\mathrm{obs}}})\) 
to a binary anomaly label 
\(y_b^t \in \{0,1\}\) at each timestep, where \(0\) denotes normal operation and \(1\) denotes anomalous behavior. 
Models are trained on a labeled dataset \(D_{\text{train}} = \{(H_b, y_b)\}\) 
and evaluated on a held-out test set \(D_{\text{test}}\), with the objective of maximizing detection performance while minimizing false alarms.

\paragraph{Metrics.}
We evaluate anomaly detection performance using precision, recall, and their harmonic mean, the F1 score. 
Precision is defined as the ratio of true positives (TP) to the sum of true positives and false positives (FP), 
while recall is the ratio of true positives to the sum of true positives and false negatives (FN). 
Formally, the F1 score is given by
\[
F1 = \frac{2 \times \text{Precision} \times \text{Recall}}{\text{Precision} + \text{Recall}}, \quad 
\text{Precision} = \frac{TP}{TP + FP}, \quad 
\text{Recall} = \frac{TP}{TP + FN}.
\]
In time-series anomaly detection, defining positive and negative samples requires care, since anomalies are typically 
labeled as contiguous incidents rather than isolated points. Following prior work \citep{gu2025argos}, 
we adopt the Event-F1 with Point Adjustment (Event-F1 PA) metric as our primary evaluation measure. 
This method treats each anomaly incident as a single detection target and considers it successfully detected 
if at least one point within the ground-truth incident is flagged. At the same time, false positives are 
penalized at the point level, which provides a balanced evaluation of both precision and recall. 
This choice ensures that models are not rewarded for overly coarse predictions and aligns with 
practical expectations in building operations, where operators require both timely and precise alarms.

\subsection*{Details of Dataset}
The assembled dataset is specifically designed to move beyond traditional binary fault detection and enable a more sophisticated diagnostic task. This section details the diagnostic targets and defines the expected output from the PILLM framework.

\paragraph{Fault Types and Intensities}

The dataset includes rich, labeled examples of various common and critical HVAC faults. The \texttt{Fault Type} provides a descriptive, human-understandable label for the specific malfunction occurring in the system. The \texttt{Fault Intensity} provides a normalized, numerical scale of the fault's severity, where a higher number indicates a more severe deviation from normal operation.

Examples of fault conditions captured in the dataset include:
\begin{itemize}
    \item \textbf{Heating Coil Leaking:} A condition where the heating coil valve is not shutting off completely, allowing hot water to leak through even when heating is not required. This leads to energy waste and potential overheating.
    \item \textbf{Damper Stuck:} An air damper is mechanically stuck at a certain position (e.g., 20\% open), preventing the system from properly regulating the mix of outdoor and recirculated air. This impacts both energy efficiency and indoor air quality.
    \item \textbf{Sensor Drift / Bias:} A temperature sensor provides consistently incorrect readings (e.g., always reporting 5°F higher than the true temperature). The system then makes incorrect control decisions based on this faulty data.
    \item \textbf{Control Logic Faults:} Such as the \texttt{Simultaneous\_Heat\_Cool} condition, where programming errors lead to inefficient and counterproductive system operation.
\end{itemize}

\paragraph{Expected PILLM Output: Generating Actionable Diagnostics}

The primary objective for the PILLM is not to predict a class label, but to generate a structured, human-readable diagnostic report. For each input "diagnostic snapshot" (i.e., a row from the dataset), the PILLM is tasked with generating a textual output that accomplishes the following:

\begin{enumerate}
    \item \textbf{Identify the Fault:} Correctly state the \texttt{Fault Type} in natural language (e.g., "The diagnosis is a stuck outdoor air damper.").
    \item \textbf{Provide Evidence:} Justify the diagnosis by referencing the physical evidence from the input data (e.g., "This is indicated because the damper position signal is fixed at 20\% while the control command is varying.").
    \item \textbf{Assess Severity:} Characterize the fault's intensity and impact (e.g., "This is a moderate-to-severe fault leading to poor ventilation and increased fan energy consumption.").
\end{enumerate}

\paragraph{Advantages Over Traditional Methods}

This diagnostic-generation task formulation offers significant advantages over conventional approaches:

\begin{itemize}
    \item \textbf{Interpretability and Trust:} Unlike a traditional classifier that outputs a cryptic label like 'Fault\_Class\_ID: 3', the PILLM's narrative output is transparent. By explaining why it reached a conclusion, it allows building operators to verify the reasoning and build trust in the system.
    \item \textbf{Actionability:} The LLM's output is directly actionable. An operator reading "inspect the outdoor air damper linkage" knows exactly what to do, whereas 'Fault\_Class\_ID: 3' would require consulting a manual.
    \item \textbf{Handling Novelty and Nuance:} By reasoning from the engineered physical features, the PILLM has the potential to describe deviations from first principles. This may allow it to characterize novel or compound faults that were not explicitly present in the training set, offering a degree of zero-shot diagnostic capability that is difficult to achieve with rigid classification models.
\end{itemize}

\subsection*{Baselines}
We compare PILLM against a diverse set of baselines, including classical deep learning models, LLM-based methods, and the recent agentic system ARGOS. Below we summarize each method included in our evaluation. 

\begin{itemize}
    \item \textbf{AnomalyTransformer}: An unsupervised model that introduces the Anomaly-Attention mechanism to detect anomalies by exploiting differences in association patterns between normal and abnormal points. This method has become a widely used benchmark in time-series anomaly detection.  

    \item \textbf{AutoRegression}: A supervised autoregressive model that applies multiple linear layers to transform input sequences into anomaly score logits. Its simplicity and efficiency make it a strong classical baseline, though it lacks adaptability to complex dependencies.  

    \item \textbf{LSTMAD}: A supervised long short-term memory (LSTM) model trained on normal data. Anomalies are detected based on statistical deviations in prediction error. It leverages temporal dependencies effectively but often struggles with generalization in highly dynamic systems.  

    \item \textbf{LLMAD}: A Large Language Model-based approach that prompts the LLM with serialized time-series data, in-context examples, and contextual information to produce anomaly predictions. While it improves interpretability compared to deep learning baselines, it suffers from non-determinism and inconsistent reproducibility.  

    \item \textbf{SigLLM}: An LLM-based method that operates in two distinct modes. In \textit{Detector mode}, the LLM predicts the next time-series values and detects anomalies by comparing them against ground truth observations. In \textit{Prompter mode}, the LLM is directly prompted with time-series data to localize anomalous indices. This design improves flexibility but often trades off precision for recall.  

    \item \textbf{ARGOS}: An agentic anomaly detection system originally developed for monitoring cloud infrastructure. ARGOS leverages LLMs to autonomously generate explainable and reproducible anomaly rules as intermediate representations, which are then deployed for efficient online detection. By combining multiple collaborative agents, ARGOS achieves explainability, reproducibility, and partial autonomy in anomaly detection. Experiments show that ARGOS outperforms prior baselines across several public and industrial datasets, highlighting the promise of LLM-driven rule-based anomaly detection. We include ARGOS as a strong state-of-the-art baseline most closely aligned with our motivation.  
\end{itemize}

\subsection*{Extra Experiment Details}

\paragraph{Hardware and Software}
All experiments were conducted on a workstation equipped with an AMD Ryzen 9 7950X 16-Core Processor and a single NVIDIA RTX 5090 GPU. The PINN framework generates anomaly detection rules as executable Python code snippets in a Python 3.12 environment, employing Google's Gemini 2.5 Flash model \citep{comanici2025gemini}.

\paragraph{Prompts}

We gather prompts used for PILLM in this section. Our prompt structure is flexible and extensible. To adapt PILLM to a new problem setting, one only needs to define its problem description, function description, and function signature.

\begin{tcolorbox}[colback=blue!5!white,
                  colframe=blue!50!black,
                  title={Prompt for population initialization}]
You are an expert in the domain of building energy, especially in heating, ventilation, and air-conditioning (HVAC). Your task is to design anomaly detection rules that can effectively detect the anomaly status of the system. 

\{ task\_description \}

Below are the input features and their descriptions for anomaly detection:

\{ input\_feature\_list\}

\{ seed\_function \}
\{ context\_template \}

Refer to the format of a trivial design above. Be very creative and give ‘{func\_name}\_v2‘. Output code only, and enclose your code in Python code and one paragraph to describe the physical hypothesis but nothing else. Format your code as a Python code string: """python ...""" and a context string: """context ...""".
\end{tcolorbox}

\begin{tcolorbox}[colback=blue!5!white,
                  colframe=blue!50!black,
                  title={System prompt for Generator LLM}]
You are an expert in the domain of building energy, especially in heating, ventilation, and air-conditioning (HVAC). Your task is to design anomaly detection rules that can effectively detect the anomaly status of the system. \{ task\_description \}. Your response outputs Python code and one paragraph to describe the physical hypothesis but nothing else. Format your code as a Python code string: """python ...""" and a context string: """context ...""".
\end{tcolorbox}

\begin{tcolorbox}[colback=blue!5!white,
                  colframe=blue!50!black,
                  title={System prompt for Reflection LLM}]
You are an expert in the domain of building energy, especially in heating, ventilation, and air-conditioning (HVAC). Your task is to provide hints for designing better anomaly detection rules. 

\{ task\_description \}

Below are the input features and their descriptions for anomaly detection:

\{ input\_feature\_list\}

You are provided with two rule versions with their physical context below, where the second version performs better than the first one.

[Worse Rules]
\{ worse\_rules \}
\{ worse\_rules\_physical\_context \}

[Better Rules]
\{ better\_rules \}
\{ better\_rules\_physical\_context \}

You respond with some hints for designing better rules and a better hypothesis as a physical context. 
\end{tcolorbox}

\begin{tcolorbox}[colback=blue!5!white,
                  colframe=blue!50!black,
                  title={System prompt for Crossover}]
You are an expert in the domain of building energy, especially in heating, ventilation, and air-conditioning (HVAC). Your task is to provide hints for designing better anomaly detection rules. 

\{ task\_description \}

Below are the input features and their descriptions for anomaly detection:

\{ input\_feature\_list\}

[Worse Rules]
\{ worse\_rules \}
\{ worse\_rules\_physical\_context \}

[Better Rules]
\{ better\_rules \}
\{ better\_rules\_physical\_context \}

[Reflection]
\{ reflection\_comments \}
\{ reflection\_context \}

[Improved Code]
Please write an improved function ‘{function\_name}\_v2‘, according to the reflection. Output code only, and enclose your code with Python code.
\end{tcolorbox}

\begin{tcolorbox}[colback=blue!5!white,
                  colframe=blue!50!black,
                  title={System prompt for Elitist Mutation}]
\{ task\_description \}

\{ input\_feature\_list\}

[Prior Reflection]
\{ reflection\_comments \}
\{ reflection\_context \}

[Code]
\{ function\_signature \}
\{ elitist\_code \}

[Improved Code]
Please write a mutated function ‘{function\_name}\_v2‘, according to the reflection. Output code only, and enclose your code with Python code.
\end{tcolorbox}

\subsection*{Extra Related Work}

Our research is positioned at the intersection of three established and one emerging field: (1) traditional Automated Fault Detection and Diagnostics (AFDD) in building systems, (2) data-driven machine learning for AFDD, (3) the drive towards physics-informed and interpretable AI, and (4) the novel application of Large Language Models (LLMs) to scientific and engineering domains.

\paragraph{Traditional and Model-Based AFDD}
The field of AFDD for buildings has a rich history, with early methods relying on physical models and expert-defined rules. These approaches can be broadly categorized into quantitative model-based methods, which compare system output to an engineering model (e.g., a simulation), and qualitative rule-based methods, which use expert knowledge to define explicit "if-then" rules for fault conditions \citep{Katipamula2005}. While highly effective for pre-defined and well-understood faults, these methods are often labor-intensive to develop, require significant domain expertise to calibrate, and can be brittle, struggling to adapt to system retrofits or novel operational conditions that fall outside their programmed logic \citep{kim2018review}.

\paragraph{Machine Learning for Fault Detection}
The increasing availability of high-frequency sensor data from Building Management Systems (BMS) has led to a surge in the application of data-driven and machine learning techniques for AFDD. These methods learn patterns directly from historical data, alleviating the need for explicit physical modeling. A wide array of techniques has been successfully applied, ranging from statistical methods like Principal Component Analysis (PCA) to supervised classifiers like Support Vector Machines (SVM) and Random Forests \citep{zhao2019artificial}. More recently, deep learning models, particularly Convolutional neural network (CNN) and Long Short-Term Memory (LSTM) networks, have shown exceptional performance in capturing the complex temporal dependencies inherent in building thermal dynamics, making them powerful tools for anomaly detection \citep{zhang2023deep}. However, while these models excel at identifying that an anomaly has occurred, they often fail to provide the necessary context to understand why.

\paragraph{The Interpretability Challenge and Physics-Informed AI}
The high performance of deep learning models often comes at the cost of interpretability. These "black box" models present a significant barrier to adoption in high-stakes environments like building operations, where trust and transparency are paramount \citep{ciobanu2024critical}. An unexplainable alert is often an ignored alert. This has fueled a growing movement towards Physics-Informed Machine Learning (PIML), which seeks to embed scientific principles into the learning process. A prominent example is the development of Physics-Informed Neural Networks (PINNs), which constrain a neural network's solution space by penalizing deviations from known physical laws, such as differential equations \citep{raissi2019physics, cuomo2022scientific}. This approach bridges the gap between data-driven flexibility and engineering rigor, leading to more robust and generalizable models. Our work builds on this philosophy, not by encoding physics into the model architecture itself, but by engineering a physics-informed feature space upon which a reasoning model can act.

\paragraph{Large Language Models as Reasoning Engines}
While originally designed for natural language tasks, the emergent capabilities of Large Language Models (LLMs) have opened new frontiers for their application in complex scientific and engineering domains. Seminal work has demonstrated that through techniques like chain-of-thought prompting, LLMs can perform multi-step reasoning, breaking down complex problems into intermediate, sequential steps in a way that mirrors human logic \citep{wei2022chain}. This ability to "think step-by-step" has unlocked performance on a wide range of arithmetic, commonsense, and symbolic reasoning tasks previously thought to be beyond the scope of language models \citep{kojima2022large}. 

This emerging body of research suggests that LLMs can function as general-purpose reasoning engines. Recent work has begun to apply these capabilities to the built environment, for example, by using LLMs to automatically design novel, physically-grounded heuristics for energy forecasting \citep{Lin2025BuildEvo}. Our PILLM framework is directly inspired by this trend. We hypothesize that an LLM's demonstrated reasoning abilities can be guided and constrained by physical principles to perform a diagnostic task that emulates a building engineer, moving beyond simple pattern recognition to generate causal, evidence-backed explanations for system faults.

\end{document}